\pdfoutput=1

\documentclass[11pt]{article}

\usepackage[final]{acl}

\usepackage{times}
\usepackage{multirow}
\usepackage{booktabs}
\usepackage{colortbl}
\usepackage{latexsym}
\usepackage{bm}
\usepackage{amssymb}
\usepackage{epigraph}

\usepackage{etoolbox}

\usepackage[T1]{fontenc}

\usepackage{algorithm}  
\usepackage{algpseudocode}  
\usepackage{enumitem}
\usepackage{amsmath}  

\usepackage[utf8]{inputenc}

\usepackage{microtype}
\usepackage{amsmath}
\usepackage{inconsolata}

\usepackage{graphicx}

\title{Establishing Trustworthy LLM Evaluation via Shortcut Neuron Analysis}

\author{Kejian Zhu\textsuperscript{1,2}\footnotemark[1], Shangqing Tu\textsuperscript{3}\footnotemark[1], Zhuoran Jin\textsuperscript{1,2} 
\textbf{Lei Hou\textsuperscript{3}}, \textbf{Juanzi Li\textsuperscript{3}\footnotemark[2]}, \textbf{Jun Zhao\textsuperscript{1,2}\footnotemark[2]} \\ \textsuperscript{1}The Key Laboratory of Cognition and Decision Intelligence for Complex Systems,\\
 Institute of Automation, Chinese Academy of Sciences, Beijing, China \\
   \textsuperscript{2}School of Artificial Intelligence, University of Chinese Academy of Sciences \textsuperscript{3}Tsinghua University\\
\texttt{zhukejian2025@ia.ac.cn, tsq22@mails.tsinghua.edu.cn} \\ 
\texttt{\{houlei, lijuanzi\} @tsinghua.edu.cn}, \texttt{jzhao@nlpr.ia.ac.cn } 
}

\definecolor{darkred}{RGB}{156, 39, 33}
\definecolor{darkblue}{RGB}{31, 90, 153}

\begin{document}
\maketitle

\renewcommand{\thefootnote}{\fnsymbol{footnote}}
\footnotetext[1]{Equal Contribution.}
\footnotetext[2]{Corresponding authors.}
\renewcommand*{\thefootnote}{\arabic{footnote}}

\begin{abstract}

The development of large language models (LLMs) depends on \textbf{trustworthy evaluation}. However, most current evaluations rely on public benchmarks, which are prone to data contamination issues that significantly compromise fairness. Previous researches have focused on constructing dynamic benchmarks to address contamination. However, continuously building new benchmarks is costly and cyclical.
In this work, we aim to tackle contamination by analyzing the mechanisms of contaminated models themselves. Through our experiments, we discover that the overestimation of contaminated models is likely due to parameters acquiring shortcut solutions in training. We further propose a novel method for identifying shortcut neurons through \textbf{comparative and causal analysis}.
Building on this, we introduce an evaluation method called \textbf{shortcut neuron patching} to suppress shortcut neurons. Experiments validate the effectiveness of our approach in mitigating contamination. Additionally, our evaluation results exhibit a strong linear correlation with MixEval (\citealp{ni2024mixeval}), a recently released trustworthy benchmark, achieving a Spearman coefficient ($\rho$) exceeding 0.95. This high correlation indicates that our method closely reveals true capabilities of the models and is trustworthy.
We conduct further experiments to demonstrate the generalizability of our method across various benchmarks and hyperparameter settings. {\textbf{Code}}: \url{https://github.com/GaryStack/Trustworthy-Evaluation}.
\end{abstract}

\section{Introduction}

Recently large language models (LLMs) have advanced rapidly, achieving remarkable results across a wide range of complex tasks (\citealp{achiam2023gpt}; \citealp{touvron2023llama}). Moreover, the open-sourcing of technology has spurred the development of numerous new models (\citealp{zhao2023survey}). In this context, evaluation has become increasingly critical, which plays a pivotal role in shaping the future trajectory of LLM development (\citealp{guo2023evaluating}).

\begin{figure}[t]
 \centering
  \includegraphics[width=0.8\columnwidth]{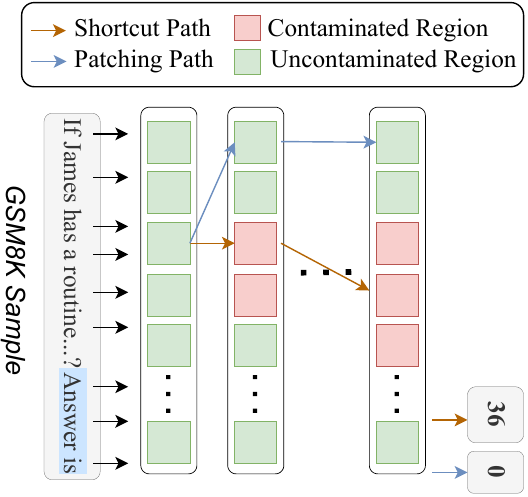}
  \caption{An example illustrating the core principle of our approach: we prevent the model from relying on shortcuts in contaminated regions to generate answers. This process restores the model’s true capabilities.}
  \label{fig:intro}
\end{figure}

We believe that trustworthiness is currently the critical aspect to enhance in evaluation, compared to evaluating the broader capabilities of LLMs (\citealp{chang2024survey}; \citealp{zhou2023don}; \citealp{yu2023kola}; \citealp{litschko2023establishing}). However, it is difficult to ensure that the large-scale and opaque training data of LLMs do not involve benchmark samples, which is called \textbf{data contamination}. Contamination can significantly affect the fairness of the evaluation (\citealp{li2024open}; \citealp{tu2024dice}). Furthermore, we highlight several critical aspects that current evaluation results overlook, compromising their trustworthiness: \textbf{A1. Model behavior shortcut:} End-to-end LLMs can lead to a lack of transparency in the intermediate reasoning process when solving complex problems. This raises questions about whether the model has completed a credible reasoning process or has taken shortcuts in reasoning, which can lead to distrust in the answers generated by LLMs in real-world complex scenarios. \textbf{A2. Input format shortcut:} The current benchmark has the drawback that the input format is fixed and differs from the way real-world inquiries are made. Models that are fine-tuned on the benchmark's input format (even the training set) have an advantage at evaluation, leading to higher scores, which is not fair.  Data contamination can lead to models being fitted to limited benchmarks, which is a key factor for aspects A1, A2 (\citealp{magar2022data}).

To address this issue, recent researches focus on developing dynamic benchmarks to mitigate contamination (\citealp{yu2023kola}; \citealp{jacovi2023stop}; \citealp{li2024latesteval}; \citealp{zhu2023dyval}). However, this strategy is resource-intensive. Besides it does not fundamentally eliminate the risk of contamination in newly released models.

To alleviate the untrustworthiness caused by contamination, it is crucial to understand the impact that contamination can have on models. We hypothesize that the untrustworthiness occurs because the model overfits the benchmark when contaminated, acquiring shortcuts for input format and reasoning. We speculate that these shortcuts in the benchmark are key to our inability to trust the model’s real-world capabilities. Through experiments, we discover a sparse set of neurons closely associated with the aforementioned shortcuts, and these shortcut neurons can be leveraged to suppress the shortcuts, as shown in \ref{exp:sparse}. Similar findings can be supported by previous studies (\citealp{golchin2023time}; \citealp{li2023estimating}; \citealp{bordt2024elephants}).

In this paper, we propose a novel method to \textbf{establish trustworthy LLM evaluation via shortcut neuron analysis}. Figure \ref{fig:intro} illustrates the principle of this approach. Recent studies have shown that transformer neurons are often closely related to specific abilities of LLMs (\citealp{geva2022transformer}; \citealp{wang2022finding}; \citealp{dai2021knowledge}). Therefore, we analyze shortcuts at the neuron level. Our method for identifying shortcut neurons is based on two key indicators: \textbf{(1) Comparative Analysis}. This involves comparing neuron activation differences between contaminated and uncontaminated models when processing the same benchmark samples. Neurons with significant activation differences are likely linked to memory shortcuts. \textbf{(2) Causal Analysis}. We compute the causal score by performing activation patching (\citealp{meng2022locating}; \citealp{Vig_Gehrmann_Belinkov_Qian_Nevo_Singer_Shieber_2020}; \citealp{zhang2023towards}) and analyzing its causal effects. A neuron is identified as a shortcut neuron if it satisfies two causal effects: (a) it restores the true scores of the contaminated model, and (b) it does not affect the normal ability of the model.

In Section \ref{exp:sparse}, we find that the shortcut neurons located above are sparse and effective, with a total of about 5000. Then we use the shortcut neuron patching method to conduct trustworthy evaluation by inhibiting shortcuts. Specifically, we will use the shortcut neuron activation of the base model with same architecture to patch models under test, so as to establish trustworthy evaluation.

To verify the effectiveness of our evaluation method, we conduct experiments on both LLaMA (\citealp{touvron2023llama}) and Mistral (\citealp{jiang2023mistral}) architectures. We fine-tune a series of contaminated and uncontaminated models. First, the accuracy of the contaminated models significantly decrease under our evaluation methodology compared to the original benchmark. This indicates that our approach effectively mitigates behavioral shortcuts in the models, enhancing their trustworthiness of black-box behavior (\textbf{A1}). 
Second, we observe that even models fine-tuned on the benchmark input format, such as those trained on the GSM8K train set, also exhibit a drop in accuracy. This suggests that our method can mitigate the input shortcuts, controlling for gains due to input format rather than model capability (\textbf{A2}).
Lastly, to verify that our method targets model shortcuts without compromising general abilities, we evaluate the patched models on math benchmark MAWPS and comprehensive benchmark MMLU. The results show that no significant accuracy changes for LLMs, indicating that our approach does not negatively impact the genuine performance of LLMs.

Additionally, we select two recently released trustworthy benchmarks, OpenMathInstruct-2 and MixEval, as reference benchmarks. MixEval is aligned with real user queries, catering to real-world model performance demands. For real-world application, we download a series of real-world models from Hugging Face. It reveals a strong linear correlation between our scores and the reference scores, with a Spearman correlation coefficient exceeding 0.95. This highlights the ability of our evaluation methodology to reliably reflect real-world model performance.
We also test the generalizability of our evaluation method across different benchmarks and hyperparameter settings, demonstrating its robustness.


In summary, our contributions are as follows:
\begin{itemize}[itemsep=0pt, leftmargin=*]
\item We are the first to analyze the neuron-level mechanism by which model's scores exceed its genuine capabilities after contamination, hypothesizing that this phenomenon is driven by shortcuts.
\item We propose a novel method for identifying neurons through comparative and causal analysis, isolating a sparse set of neurons closely associated with shortcut reasoning.
\item We introduce shortcut neuron patching method to enable more trustworthy evaluation by suppressing shortcuts for both input format and behavior.
\end{itemize}

\section{Related Work}

\subsection{Data Contamination}

Data contamination refers to the inclusion of benchmark data in the training phase of machine learning models, resulting in artificially inflated benchmark scores (\citealp{magar2022data}). This issue is particularly pronounced in the era of LLMs, which are trained on massive corpus. (\citealp{brown2020language}, \citealp{magar2022data}). Such contamination raises significant concerns about the validity of benchmarking studies and the generalizability of LLMs (\citealp{sainz2023nlp}; \citealp{xu2024benchmarking}).

\subsection{Mitigate Contamination}

To mitigate the impact of contamination for trustworthy evaluation, recent studies have approached this challenge by proposing dynamic benchmark construction and updating protocols to minimize overlap with pre-training data (\citealp{zhu2023dyval}, \citealp{zhu2023clean}). On the one hand, \citealp{yu2023kola} introduce a dynamic evaluation framework, reshaping the static nature of benchmarks. On the other hand, benchmark data encryption and label protection have also been suggested as strategies to prevent contamination (\citealp{jacovi2023stop}). Besides, there is also a work sampling on the distribution of model outputs and removing outputs that are most likely to be affected by contamination to conduct a trustworthy evaluation (\citealp{dong2024generalization}).

\subsection{Transformer Neuron}

In transformer-based LLMs, each layer $l$ consists of a multi-head attention mechanism followed by a feed-forward network (FFN), which is a multi-layer perceptron (MLP) (\citealp{vaswani2017attention}). The FFN is defined as:
\begin{equation}
    \text{FFN}(x) = \sigma(x K^\top + b_1) V + b_2
\end{equation}
where $x$ is the input to the layer, $K$ and $V$ are the weight matrices, \( b_1 \) and \( b_2 \) are the bias terms, and \( \sigma \) is a non-linear activation function.

The neuron in LLM specifically refers to activation before down projection in MLP, which has been shown to be critical for information processing (\citealp{geva2022transformer}). The activation of a neuron \( v_j^l \) is determined by its corresponding activation coefficient \( m_{ij}^l \), which is calculated as:
\begin{equation}
    m_{ij}^l = \sigma(x_i^l \cdot k_j^l)
\end{equation}
where \( x_i^l \) is the representation of token \( x_i \) at layer \( l \), \( k_j^l \) is the \( j \)-th row of \( K^l \) in the MLP.

Previous excellent work has found that transformer neurons are correlated with certain aspects of LLM capabilities (\citealp{geva2020transformer}). Therefore, recent works have studied the mechanism of LLM through LLM neurons, and various types of neurons have been discovered. For example, (\citealp{dai2021knowledge}) identified "knowledge neurons" that appear to store factual knowledge, while (\citealp{wang2022finding}) discovered "skill neurons" associated with specific linguistic skills. There are also concept neurons (\citealp{geva2022transformer}), safety neurons (\citealp{chen2024finding}), etc.  Recent works usually project neuron representations to the vocabulary space to study the mechanism and meaning of neurons (\citealp{geva2022transformer}), and verify the function of neurons through causal analysis (\citealp{ghandeharioun2024patchscope}, \citealp{gurnee2024universal}). However, they usually focus on theoretical analysis at the neuron mechanism level, but lack practical application scenarios. In this paper, we try to find shortcut neurons for contaminated models, which may be the main reason why the model scores on the contaminated benchmark are artificially high.


\begin{figure*}[t] 
    \centering
  \includegraphics[width=\linewidth]{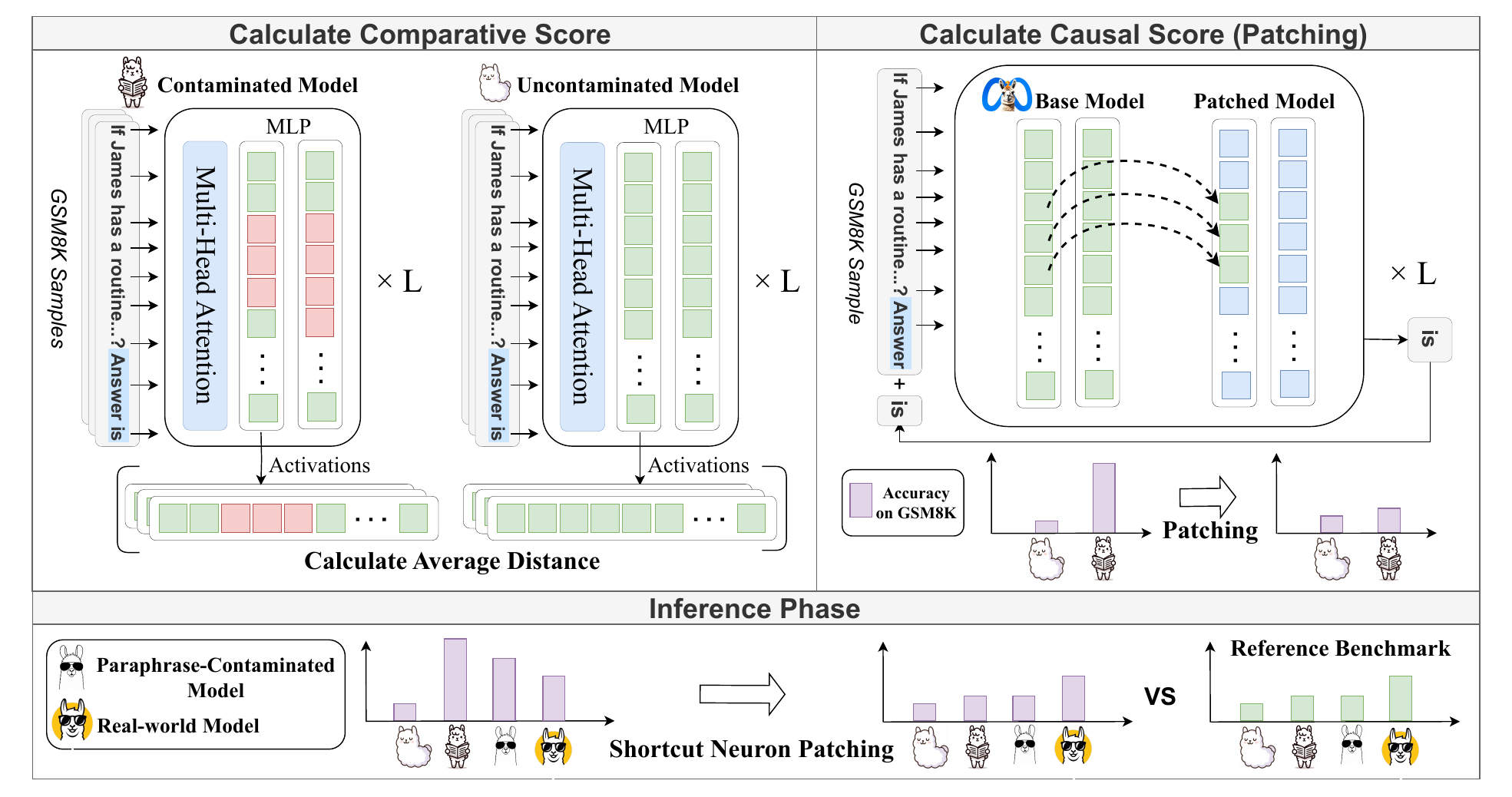}
  \caption{The overview of our method. We employ neuron analysis to identify regions within the model that may be overestimating its capabilities due to shortcuts. We calculate comparative and causal scores to find shortcut neurons. The former highlights the areas where there is the greatest divergence between parameters of contaminated and uncontaminated models. The latter is derived from neuron patching analysis to assess its causal impact. Subsequently, we use the located shortcut neurons to patch various models under test to obtain trustworthy evaluation results.}
  \label{fig:methodology}
\end{figure*}

\section{Methodology}
\label{method}

In this section, we propose our methodology for trustworthy evaluation. We restore the true capability of the contaminated model by suppressing the impact of shortcuts.

\subsection{Overview}
\label{methodoverview}

Most previous works on trustworthy evaluation focus on constructing uncontaminated benchmarks. For example, some efforts create benchmarks using the most recent texts (\citealp{li2024latesteval}), while others develop dynamic benchmarks (\citealp{yu2023kola}, \citealp{zhu2023dyval}). However, since LLMs are continuously updated, ensuring the timeliness of these benchmarks remains a significant challenge.

Unlike previous works, we turn attention on the inner mechanism of models to study the origin of the overestimation.  We hypothesize that contamination provides the model with shortcut solution, leading to an overestimation of its abilities. We found that some neurons in the contaminated models are associated with their high scores on contaminated benchmark. We refer to these as \textbf{shortcut neurons}. Our proposed method identifies and patches these shortcut neurons to suppress shortcuts within the model, mitigate the effects of contamination.

As illustrated in Figure \ref{fig:methodology}, our approach comprises two primary phases. (1) We locate shortcut neurons using contrasting distance (\ref{contrastinglocate}) and causal analysis (\ref{causallocate}), as detailed in Section \ref{causallocate}. (2) We apply dynamic patching technique to validate the causal effects of shortcut neuron and evaluate their effectiveness in mitigating the impact of data contamination, as discussed in Section \ref{patch}.

\subsection{Locate: Comparative Analysis}
\label{contrastinglocate}


Before locating the shortcut neurons of a model architecture $\mathcal{M}$, we need to fine-tune the vanilla model $\mathcal{M}_\text{0}$ of this architecture to get contaminated and uncontaminated models. For convenience, we denote the contaminated and uncontaminated model as $\mathcal{M}_\text{con}$ and $\mathcal{M}_\text{un}$ respectively. For a given input token $\bm x_t$, we represent the activation of the $i^{th}$ neuron in layer $l$-th of $\mathcal{M}$  as $\bm{a}^l_i (\bm x_t|\mathcal{M})\in\mathbb{R}$. Given a prompt $\bm X=\langle x_0, x_1, ..., x_T \rangle$, the activation representation of a given neuron can be computed either by the average activation on all tokens ($\bm{a} = \frac{1}{T}\sum_{t=1}^{T} \bm{a}_t$), or by using activation on the last token ($\bm{a} = \bm{a}_T$). We adopt the last token's activation because it more effectively captures the overall activation feature of the entire prompt (\citealp{zhao2024explainability}, \citealp{wang2023improving}).
Let $\mathcal{D}$ denote the dataset with data contamination. We define the comparison score for the $i^\text{th}$ neuron in the $l^\text{th}$ layer on $\mathcal{D}$ as the root mean square of the differences between the activations of models $\mathcal{M}\text{con}$ and $\mathcal{M}\text{un}$ during the generation process:
\begin{equation}
\begin{aligned}
    &\bm{S}^l_i (\mathcal{M}, \mathcal{D}) = \\ 
     &\sqrt{\frac{\sum_{x \in \mathcal{D}} \left(\bm{a}^l_i (\bm x_T|\mathcal{M}_\text{con})-\bm{a}^l_i (\bm x_T|\mathcal{M}_\text{un})\right)^{2}}{|\mathcal{D}|}}\\
\end{aligned}
\end{equation}

\subsection{Locate: Causal Analysis}
\label{causallocate}

\textbf{Activation patching.}
Activation patching (\citealp{vig2020investigating}) is the most prevalent method for evaluating causal effects of neurons on LLMs. Traditionally, this method has been applied to short output tasks, where the focus is on assessing how much we can restore the probability of predicting the next correct token on the corrupted input with activation patching. However, contamination occurs in various task scenarios, such as mathematics (\citealp{tu2024dice}), coding(\citealp{matton2024leakage}), etc. The outputs of these scenarios are open ended, so dynamic patching is required. In dynamic patch, we'll use the activation of the patching model's neurons to replace the activation of the corresponding neurons in the patched model in generation process. In detail: (1) Run patching model $\mathcal{M}_\text{patching}$ on current prompt $\bm 
 X_\text{t}$ and cache activations of given neurons; (2) Run 
$\mathcal{M}_\text{patched}$ on the same prompt $\bm 
 X_\text{t}$ with the activation of given neurons replaced by cached activation while the other neurons keep unchanged; (3) Predict next token $\bm x_\text{t}$ and append it to current prompt for a new one $\bm X_\text{t+1} = \bm X_\text{t} + \bm x_\text{t}$. Repeat above steps until generation process finished.

\textbf{Calculate Causal Score of Neurons.}
A neuron that is responsible for contamination or memorization should have two important features: (1) It has a significant impact on the performance of the contaminated model. (2) It has as little impact on the model's own capabilities as possible, which can also be characterized as having little impact on the performance of the uncontaminated model. Based on this assumption, we use dynamic patching method to calculate the causal score of each neuron. Similar to \ref{contrastinglocate}, we use the vanilla model $\mathcal{M}_\text{0}$ as the patching model, while $\mathcal{M}_\text{con}$ and $\mathcal{M}_\text{un}$ as the patched models. Assume that the prompt dataset with data contamination is $\mathcal{D}$, we define causal score of investigated neurons set $\mathcal{N}$ is:
\begin{equation}
    \begin{aligned}
    \bm{C}_\mathcal{N} = & a(\mathcal{M}_\text{con})-a_\text{patch}(\mathcal{M}_\text{con}|\mathcal{M}_\text{0}) \\
    &  + 1-(a(\mathcal{M}_\text{un})-a_\text{patch}(\mathcal{M}_\text{un}|\mathcal{M}_\text{0}))
    \end{aligned}
\end{equation}
where $a(\mathcal{M}_\text{con})$ represent the accuracy of model $\mathcal{M_\text{con}}$ on $\mathcal{D}$; $a_\text{patch}(\mathcal{M}_\text{con}|\mathcal{M})$ represent the accuracy after patched by guided model $\mathcal{M}$. $a(\mathcal{M}_\text{un})$ and $a_\text{patch}(\mathcal{M}_\text{un}|\mathcal{M})$ have similar meaning. In the above formula, if the performance of $\mathcal{M_\text{con}}$ is worse after the patch, $\bm{C}_\mathcal{N}$ is higher, and if the performance of $\mathcal{M}_\text{un}$ is worse, $\bm{C}_\mathcal{N}$ is lower.


\subsection{Trustworthy Evaluation}
\label{patch}

We aim to achieve more trustworthy evaluation results by addressing two critical aspects of trustworthiness (\textbf{A1, A2}). Our goal is to suppress the supernormal performance brought about by behavior and input shortcuts without affecting the model's true capabilities. In the Locate section, we identified neurons associated with model shortcuts, and next, we will propose a shortcut neuron patching evaluation framework.

We replace the activations of shortcut neurons in model to be evaluated $\mathcal{M}_{e}$ with those in base model $\mathcal{M}_\text{0}$, so as to suppress the contaminated model from shortcut reasoning. This enables us to mitigate the adverse effects of data contamination to a certain extent and restore the true performance on the contaminated benchmark. Specifically, a contaminated model $\mathcal{M_\text{con}}$ that is fine-tuned from base model $\mathcal{M_\text{0}}$ on the contaminated benchmark should perform at a similar level to $\mathcal{M_\text{0}}$ after being patched; while a uncontaminated model $\mathcal{M_\text{un}}$ that is fine-tuned on an irrelevant dataset should have almost no effect on the performance after being patched. This allows trustworthy evaluation to be achieved in the presence of contamination.

By leveraging this dual-phase methodology, we aim to enhance the robustness of LLM evaluation against data contamination and contribute to the development of trustworthy evaluation practices.

\begin{table*}
  \centering
  \fontsize{9pt}{12pt}\selectfont 
  \begin{tabular}{lccc}
    \hline
    \textbf{Label}           & \textbf{Benchmark Samples} & \textbf{Occurrences} & \textbf{Base Models} \\
    \hline
    contaminated       & \{GSM-i, GSM-i-Syn\}           &  \{1,5\}       & \multirow{2}{*}{\{LLaMA2-7B, Mistral-7B-v0.2\}}                    \\
    uncontaminated     & \{GSM8K Train, MATH, MATH-Syn\}          &   \{1\}            \\
    \hline
  \end{tabular}
  \caption{\label{finetuned-models}
    The models needed in the trustworthy evaluation experiment are all fine-tuned from the given basic models, simulating a variety of contaminated and uncontaminated models in the real world.
  }
\end{table*}
\begin{table*}
  \centering
  \setlength\tabcolsep{5pt}
\begin{tabular}{lcccccccc}
\toprule
 & \multicolumn{4}{c}{\textbf{LLaMA2-7B}} & \multicolumn{4}{c}{\textbf{Mistral-7B}} \\
 \cmidrule(lr){2-5} \cmidrule(lr){6-9}
 & \textbf{Ref Acc} & \textbf{Ori.} & \textbf{TE} & \textbf{$ \Delta_\text{acc} $} & \textbf{Ref Acc} & \textbf{Ori.} & \textbf{TE} & \textbf{$\Delta_\text{acc}$} \\
\midrule
 Vanilla & 16.7 & 18.5 & 18.5 & - & 31.8 & 40.0 & 40.0 & - \\
~+GSM-i & 26.7 & 40.5 & 27.0 & \cellcolor[HTML]{FDB594} \textcolor{darkred}{-13.5} & 35.2 & 58.5 & 42.0 & \cellcolor[HTML]{FCAB84} \textcolor{darkred}{-16.5} \\
~+GSM-i-Syn & 23.3 & 33.4 & 20.5 & \cellcolor[HTML]{FED9C7} \textcolor{darkred}{-12.9} & 36.0 & 48.6 & 41.5 & \cellcolor[HTML]{FEE5D9}\textcolor{darkred}{-7.1} \\
~+5$\times$GSM-i & 23.7 & 80.0 & 30.2 & \cellcolor[HTML]{FC8D59} \textcolor{darkred}{-49.8} & 39.5 & 88.7 & 45.6 & \cellcolor[HTML]{FC8D59} \textcolor{darkred}{-43.1} \\
~+5$\times$GSM-i-Syn & 24.7 & 46.5 & 26.8 & \cellcolor[HTML]{FCAB84} \textcolor{darkred}{-19.7} & 38.3 & 56.1 & 43.3 & \cellcolor[HTML]{FDB594} \textcolor{darkred}{-12.8} \\
      \midrule
~+OpenOrca & 21.0 & 20.2 & 21.5 & \cellcolor[HTML]{C3DCEC}\textcolor{darkblue}{+1.3} & 36.5 & 42.5 & 43.0 & \cellcolor[HTML]{DCEBF4}\textcolor{darkblue}{+0.5} \\
~+GSM8K Train & 24.6 & 35.0 & 28.5 & \cellcolor[HTML]{FED9C7}\textcolor{darkred}{-6.5} & 42.8 & 49.6 & 45.3 & \cellcolor[HTML]{FEE5D9}\textcolor{darkred}{-4.3} \\
~+MATH & 20.6 & 19.5 & 19.0 & \cellcolor[HTML]{FEF6F1}\textcolor{darkred}{-0.5} & 30.5 & 39.5 & 38.2 & \cellcolor[HTML]{FEF3ED}\textcolor{darkred}{-1.3} \\
~+MATH-Syn & 22.1 & 20.3 & 20.5 & \cellcolor[HTML]{DCEBF4}\textcolor{darkblue}{+0.2} & 32.5 & 41.3 & 42.0 & \cellcolor[HTML]{CCE1EE}\textcolor{darkblue}{+0.7} \\

\bottomrule
\end{tabular}
\caption{\label{trustworthy evaluation results}
    Trustworthy evaluation in the presence of contamination. Ori.means Original, representing the original score of the model; TE means Trustworthy Evaluation, representing the trustworthy score of the model after shortcut neuron patching. 5$\times \mathcal{D}$ represents that data of $\mathcal{D}$ occurs 5 times in training phase. For Ref Acc, we selected OpenMathInstruct-2 (\citealp{toshniwal2024openmathinstruct}) dataset as the reference standard. $ \Delta_\text{acc} $ represents TE score minus Ori. score. Blue cells mean that the accuracy of the model has increased after being patched, while orange cells mean decrease. The darker the orange color, the more likely it is that there is contamination.
  }
\end{table*}

\section{Experiment}
\label{Exp}

\subsection{Experimental Setup}
\label{sec:exp_setup}

\textbf{Datasets.} We use mathematical reasoning benchmarks as contaminated dataset. Specifically, we conduct experiments on GSM8K (\citealp{cobbe2021training}), MATH (\citealp{hendrycks2021measuring}), SVAMP (\citealp{patel2021nlp}), ASDiv (\citealp{miao2021diverse}) and MAWPS (\citealp{koncel2016mawps}). Because these datasets are all used to evaluate the mathematical reasoning ability of models and have similar distributions (\citealp{gou2023tora}). 

\noindent
\textbf{Base Architecture.} To test the effectiveness of our method, we select two LLM frameworks with high recognition: LLaMA2-7b (\citealp{touvron2023llama}), Mistral-7b-v0.2 (\citealp{jiang2023mistral}). 

\noindent
\textbf{Models.} Following prior work (\citealp{dekoninck2024evading}), we simulate contamination by fine-tuning LLaMA2-7B and Mistral-7B-v0.2 to create contaminated and uncontaminated models, which are detailed in Table \ref{finetuned-models}. GSM-i represents 50\% of the GSM8K test set, comprising a total of 657 samples. $\mathcal{D}$-Syn is generated by paraphrasing the original questions and answers from benchmark $\mathcal{D}$ (ensuring correctness) using GPT-4 (\citealp{achiam2023gpt}). To ensure uniformity, benchmark samples are mixed with OpenOrca instruction data (\citealp{lian2023openorca}), resulting in a training dataset of 25,000 samples.

\noindent
\textbf{Implementation Details.} For the hyperparameters that are used for sampling strategies of LLMs' decoding, we set \textit{temperature} to 1, \textit{top-p} to 1 and \textit{top-k} to 50 throughout the experiments. Due to the large number of neurons in LLMs, we select 512 adjacent neurons as a group to calculate the causal effect as a whole during the locate process.

\begin{figure}
  \includegraphics[width=\columnwidth]{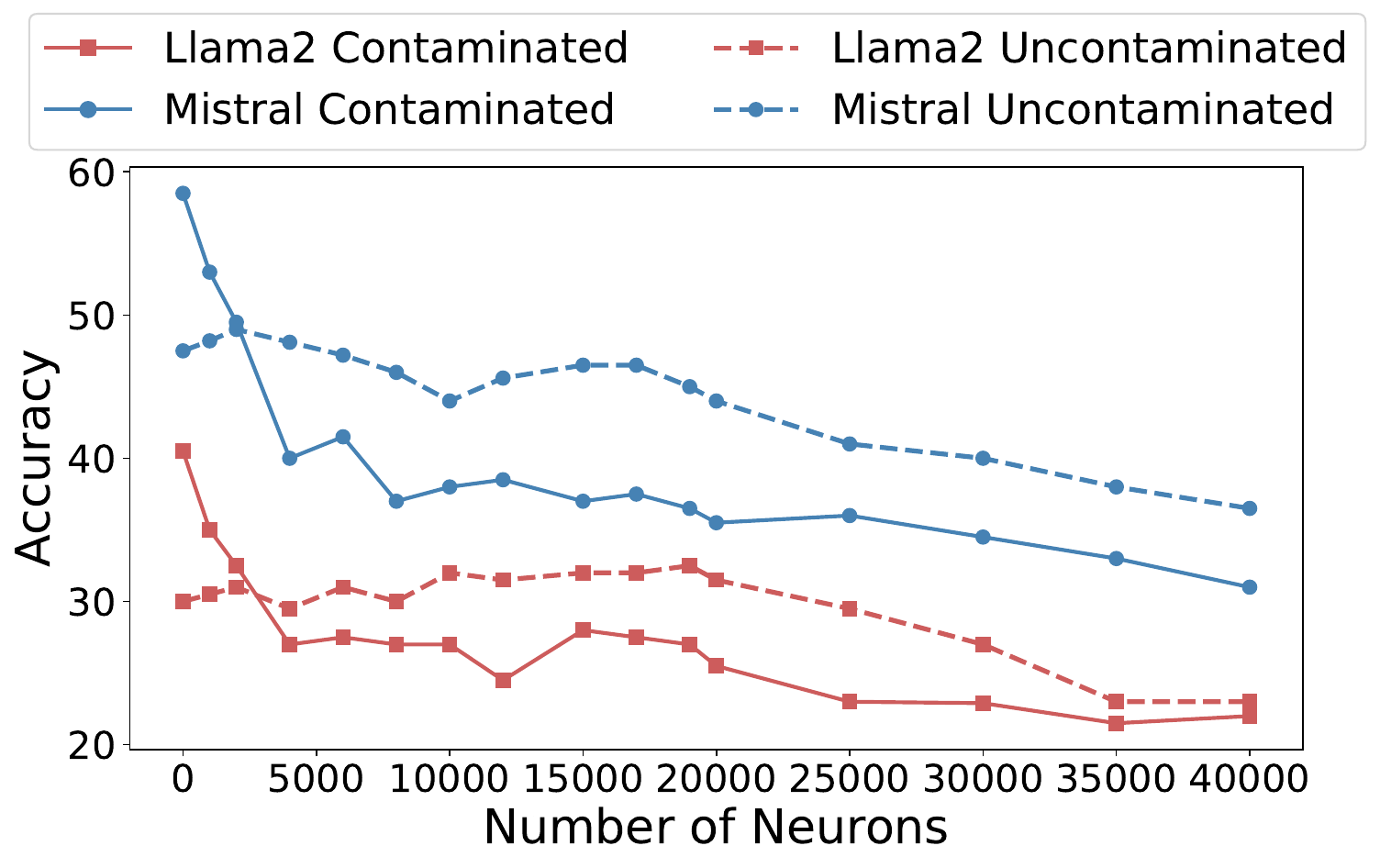}
  \caption{The performance of the contaminated and uncontaminated models changes as the number of neurons in the patch increases, using experiments with located shortcut neurons and random neurons, respectively.}
  \label{fig:sparse}
\end{figure}

\subsection{Shortcut Neurons Are Sparse}
\label{exp:sparse}
In the previous section, we introduce how to calculate the shortcut score of each neuron. However, how many of the top neurons ranked by score are related to memory shortcuts still need to be explored. Because if too many neurons unrelated to contamination are patched, it may affect the performance of both the contaminated model and the uncontaminated model. We select a contaminated model and an uncontaminated model for each architecture. Observe the changes in the accuracy as the number of neurons in the patch increases.

Figure \ref{fig:sparse} shows that after 5,000 neurons were patched, the accuracy of the contaminated model has roughly reached the same level as the uncontaminated model, and the accuracy of the uncontaminated model has changed very little. After 20,000 neurons are patched, the accuracy of both models begins to decline. This result shows that the first 5,000 neurons have a good effect on alleviating model contamination. 5,000 neurons only account for 1.4\% of Llama2-7B neurons and 1.1\% of Mistral-7B neurons, indicating that shortcut neurons are sparse. 


\begin{figure*}[t] 
  \includegraphics[width=0.48\linewidth]{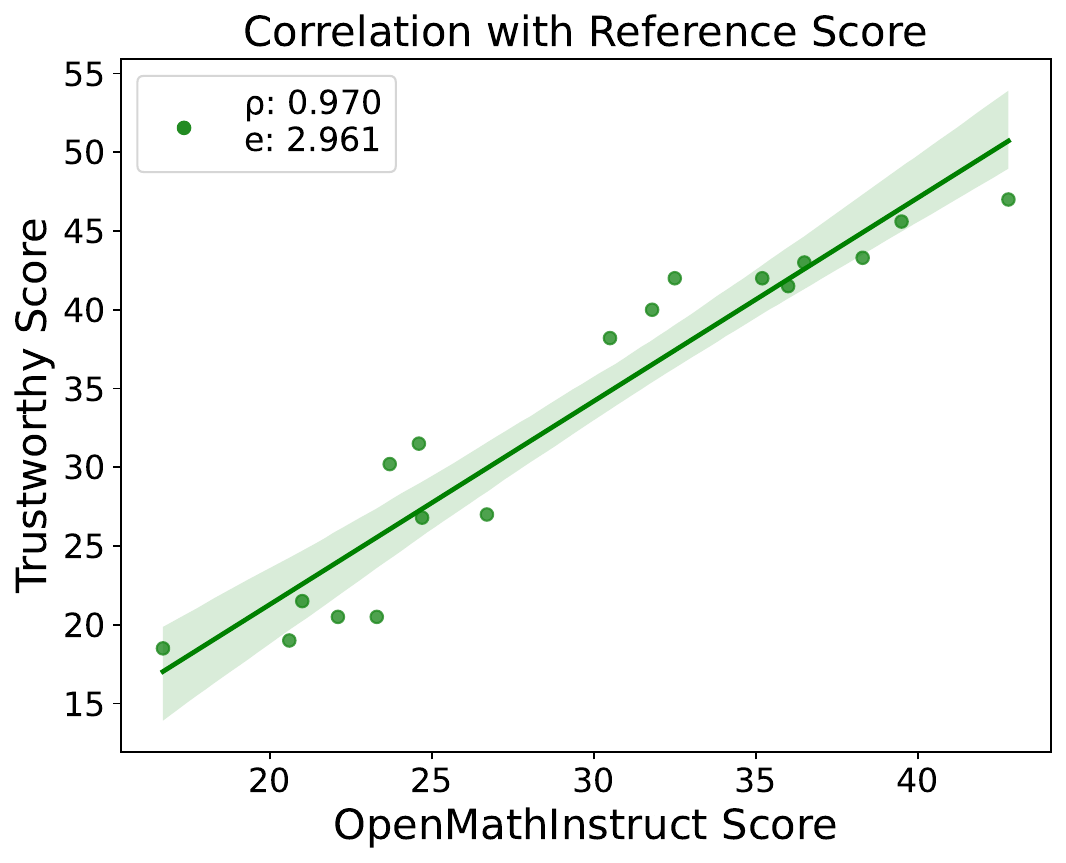} \hfill
  \includegraphics[width=0.49\linewidth]{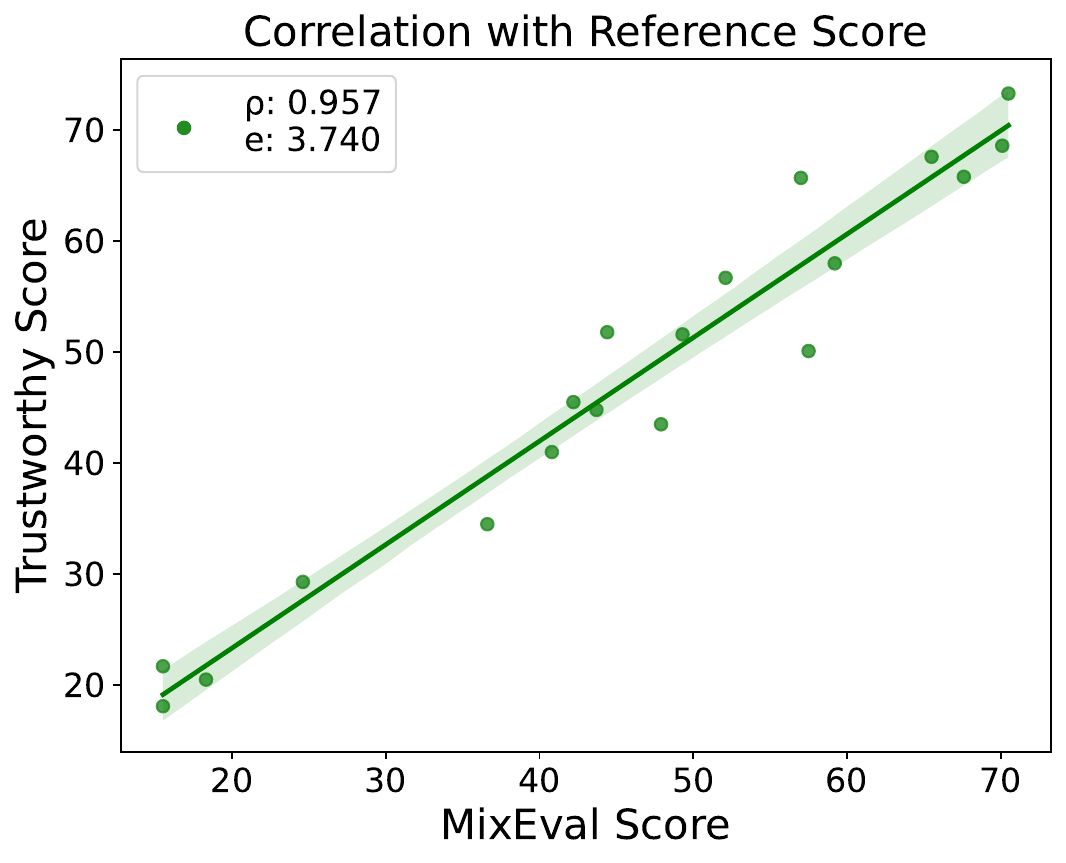}
  \caption {Correlation between the trustworthy evaluation scores obtained by our method and the reference scores in simulation and real-world settings. We choose OpenMathInstruct-2 ((\citealp{toshniwal2024openmathinstruct})) and MixEval (\citealp{ni2024mixeval}) as reference for simulation and real-world evaluation respectively. $\rho$ and $e$ denote the Spearman’s ranking correlation and the root mean square error (RMSE) of the linear correlation respectively. }
    \label{fig:correlation_ref}
\end{figure*}

\subsection{Results of Trustworthy Evaluation}
\label{exp:Main Results}

In this section, we will present the results of evaluation and analyze the effectiveness in addressing the two trustworthiness factors previously discussed. Following the finding above, shortcut neurons are selected as the top 5000 neurons.

\noindent
\textbf{Trustworthiness for Model Behavior. } \label{exp:Model Behavior} Ensuring that the black box model gets the answer through multi-hop reasoning rather than shortcuts from contamination is the key to trustworthiness. To verify this, we will use fine-tuned models to conduct simulation experiment. Specifically, we select GSM8K as example. For a model, we test its original accuracy on GSM-i and accuracy after patching.

The results of the simulation settings are presented in Table \ref{trustworthy evaluation results}. Notably, the performance of contaminated models decreases significantly after patching, with an average drop of 37\%, highlighting the effectiveness of shortcut neuron patching in mitigating contamination. Meanwhile, the accuracy of the uncontaminated model changes by 3\% on average, demonstrating that our method has minimal impact on the reasoning ability of models. It proves that we can effectively suppress model shortcuts and improve the credibility of model behavior. We can also improve the transparency of the intermediate process of model behavior and ensure that the source of the score is the model's ability. Furthermore, we select the OpenMathInstruct-2 math problem dataset (\citealp{toshniwal2024openmathinstruct}) recently released by NVIDIA as an uncontaminated benchmark as reference. Figure \ref{fig:correlation_ref} illustrates a strong positive correlation between our score and the reference score, with a Spearman correlation coefficient $\rho$ of 0.970. This shows that the scores obtained by patching can achieve a more trustworthy evaluation by avoiding shortcuts that contamination brings.

\noindent
\textbf{Trustworthiness for Model Input. }
\label{exp:Model Input} The input format of the benchmark is fixed and may differ from real-world user queries. The model may fit this input method by training on the benchmark (including the training set with the same format), which will cause the score to exceed the actual level. We call this overestimation an input shortcut. Table \ref{trustworthy evaluation results} also shows the suppression of input shortcuts by our method. It can be observed that the accuracy of the contaminated model has decreased due to fine-tuning on input formats. Specifically, the uncontaminated model fine-tuned on the GSM8K training set, which fits the same format, has also experienced a decline in accuracy.

\begin{table}
    
  \setlength\tabcolsep{5pt}
\begin{tabular}{lcccc}
\toprule
 & \multicolumn{2}{c}{\textbf{MAWPS}} & \multicolumn{2}{c}{\textbf{MMLU}} \\
 \cmidrule(lr){2-3} \cmidrule(lr){4-5}
 & \textbf{Ori.} & \textbf{TE} & \textbf{Ori.} & \textbf{TE} \\
\midrule
Vanilla LLaMA & 29.1 & 29.1 & 45.9 & 45.9  \\
~+GSM-i & 39.8 & 37.5 & 53.2 & 51.5  \\
~+GSM-i-Syn & 37.9 & 42.1 & 50.6 & 51.0 \\
~+5$\times$GSM-i & 29.2 & 25.5 & 48.1 & 46.5 \\
~+5$\times$GSM-i-Syn & 24.4 & 24.5 & 43.8 & 42.5 \\
~+OpenOrca & 23.2 & 28.6 & 59.7 & 58.6 \\
~+GSM8K Train & 39.9 & 45.2 & 51.8 & 53.4 \\
~+MATH & 21.5 & 18.5 & 40.6 & 41.0 \\
\bottomrule
\end{tabular}
\caption{\label{side effects}
    The scores of different models on elementary school math problems and reasoning datasets before and after patching. We choose MAWPS (\citealp{koncel2016mawps}) and MMLU (\citealp{hendrycks2020measuring}) to analyze the reasoning ability of the model and it will not be affected by the shortcut neuron being patched.
  }
\end{table}

\noindent
\textbf{Is There a Side Effect?} \label{exp:Side Effect} We further investigate whether our method would impact the model's normal capabilities, ensuring that it only suppresses unfair shortcuts that the model takes during the evaluation cycle (data origin, input, and inference behavior). Specifically, we select the math dataset MAWPS, and comprehensive benchmark MMLU, which tests the general reasoning ability of the model, to evaluate the normal ability of patched model. We find that although the activation values of the 5,000 shortcut neurons of these models were changed, it do not have a significant impact on their scores, as shown in Table \ref{side effects}.

\noindent
\textbf{Real World Application.}\label{exp:real-world} For the Mistral-7B and LLaMA2-7B frameworks, we select several real-world LLMs available on Hugging Face for evaluation. Detailed information about the models and their results is provided in Appendix~\ref{appendix: real-world}.
We select the math part of MixEval (\citealp{ni2024mixeval}), a recent and highly recognized evaluation work, as reference benchmark. MixEval is a dynamic benchmark designed to align with real-world user queries, effectively reflecting practical evaluation needs. Figure \ref{fig:correlation_ref} illustrates the relationship between our evaluation scores and the MixEval scores. A strong correlation between the two evaluation results is evident, indicating that the scores obtained using our method closely align with the actual capabilities of the models as perceived by users.

\begin{table}
  \setlength\tabcolsep{4pt}
\begin{tabular}{lccc}
\toprule
 & \multicolumn{3}{c}{\textbf{MAWPS}} \\
 \cmidrule(lr){2-4} 
 & \textbf{Ori.} & \textbf{TE} & \textbf{Ref.} \\
\midrule
 Vanilla & 29.1 & 29.1 & 16.7 \\
~+MAWPS & 46.5 & \cellcolor[HTML]{FCAB84} 33.0(\textcolor{darkred}{-13.5}) & 25.7 \\
~+MAWPS-Syn & 39.4 & \cellcolor[HTML]{FDB594} 28.1(\textcolor{darkred}{-11.3}) & 21.3 \\
~+5$\times$MAWPS & 83.2 & \cellcolor[HTML]{FC8D59} 38.5(\textcolor{darkred}{-44.7}) & 28.5 \\
~+5$\times$MAWPS-Syn & 41.7 & \cellcolor[HTML]{FED9C7} 32.5(\textcolor{darkred}{-9.2}) & 23.1 \\
      \midrule
~+OpenOrca & 32.0 & \cellcolor[HTML]{A4CAE1} 33.6(\textcolor{darkblue}{+1.6}) & 21.0\\
~+SVAMP & 37.8 & \cellcolor[HTML]{FEF6F1} 37.0(\textcolor{darkred}{-0.8}) & 26.2 \\
~+ASDiv & 34.5 & \cellcolor[HTML]{C3DCEC} 35.8(\textcolor{darkblue}{+1.3}) & 23.5 \\

\bottomrule
\end{tabular}
\caption{\label{tab:generalization}
    Use the shortcut neuron located before to perform trustworthy evaluation on other mathematical reasoning datasets. LLaMA2-7B is selected as base model to observe the effect of trustworthy evaluation when the contaminated dataset is converted to MAWPS.
  }
\end{table}

\subsection{Generalization}
\label{exp:generalization}

\textbf{Generalization on Different Datasets. }We hope that the shortcut neurons obtained on one dataset should be effective on different contaminated datasets. So we set the contaminated datasets to MAWPS and MATH to observe whether the shortcut neurons located for GSM8K still work. We also fine-tune a series of models (shown in Appendix~\ref{appendix:generalization on benchmarks}) and find that under the contaminated settings of MAWPS and MATH, this batch of shortcut neurons can also help us achieve the purpose of trustworthy evaluation, as shown in Table \ref{tab:generalization}.

\begin{figure}
    \centering
  \includegraphics[width=0.93\columnwidth]{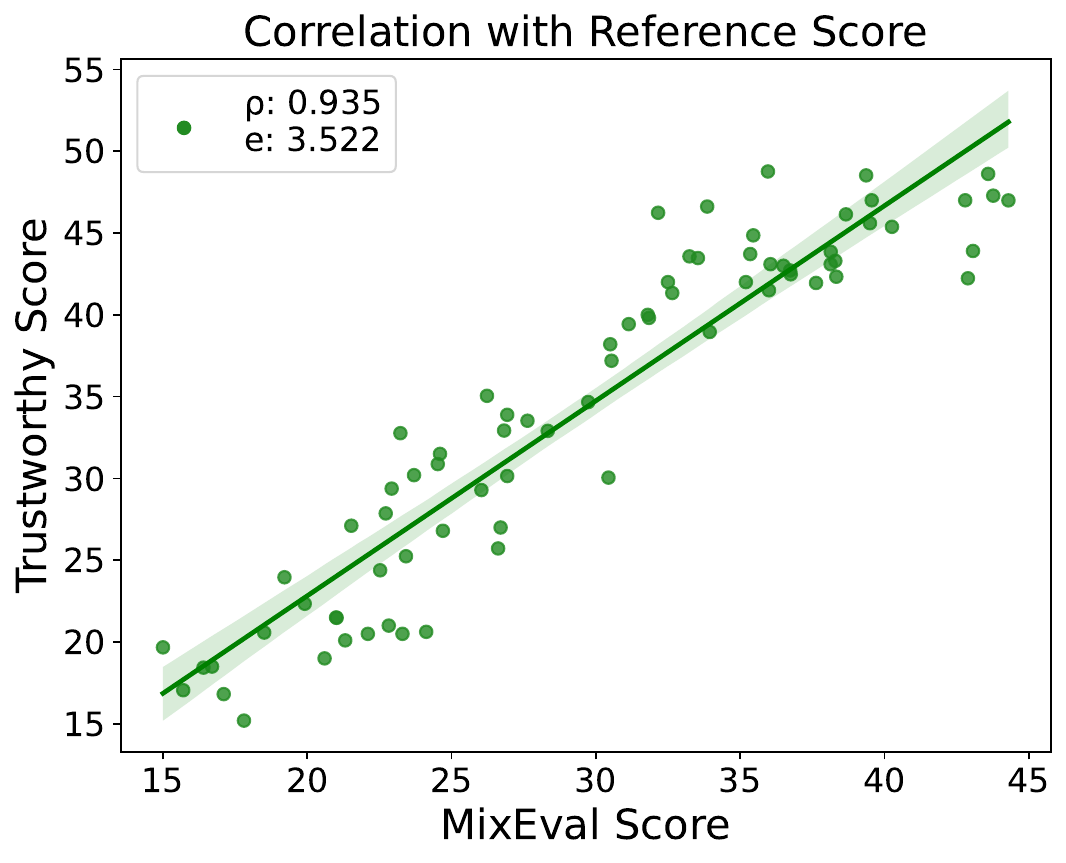}
  \caption{A figure to demonstrate the generalizability of shortcut neuron. Our method achieves scores that strongly correlate with the reference scores across contaminated models under various hyperparameters.}
  \label{fig:generalization}
\end{figure}

\textbf{Generalization across Various Hyperparameters. }We also discuss whether our method can still address the two aspects of trustworthy evaluation (\textbf{A1, A2}) when the model's training hyperparameters are changed. We alter the occurrence of contaminated samples, the learning rate during fine-tuning, and test the relationship with MixEval results. From Figure \ref{fig:generalization}, it can be observed that our results still align with the model capabilities provided by real-world users under different hyperparameters, demonstrating robustness.

\section{Conclusion}
\label{conclusion}
In this paper, we present a novel trustworthy evaluation method. Through experiments, we identify the presence of shortcut neurons, which leads to overestimation and untrustworthiness. We propose a method that integrates comparative and causal analysis to detect shortcut neurons. 
Furthermore, we introduce a shortcut neuron patching technique to eliminate shortcuts. Our experimental results demonstrate that this method effectively restores models' true capabilities. Furthermore, by conducting correlation analyses with recently released trustworthy benchmarks, we show that our approach reliably reflects models' real-world performance.


\section*{Limitations}
\label{sec:bibtex}
Although we have done our best to do a lot of experiments, some aspects are still not covered:

(1) Due to the limitation of computing resources, we only discussed two frameworks (e.g. LLaMA2-7B, Mistral-7B-v0.2). In the future, we will expand our research to more frameworks.

(2) In simulation experiments, we used the full parameter fine-tuning method to obtain the models, instead of using pre-training. Here we assume that base models are uncontaminated, but in fact, even base models cannot completely eliminate the suspicion of contamination. However training a clean model from scratch is very expensive.

(3) Our experiments are mainly conducted on mathematical reasoning benchmarks, which we believe are the most representative of data contamination. In the future, we will apply the shortcut neuron patching method to other broader benchmarks to contribute to LLM evaluation.

(4) We found that there are large differences in shortcut neurons under different architectures, which may affect the generalization of our method. We will further study this issue in the future.

\section*{Ethics Statement}

Our work has explored some mechanisms in the complex LLM network. However, good mechanism research methods may be used to influence LLM's autonomous decision-making in high-risk scenarios or even generate harmful outputs (avoiding safety alignment). Understanding the mechanism of the model does not mean that the model can be fully trusted. The safety of technological development must be guaranteed from an ethical perspective. Besides, we used AI assistants to check grammar and polish the text of the paper. But we carefully checked and made sure that the AI assistant did not change the original meaning of the article. For open-accessible datasets used, we have checked their licenses.

\section*{Acknowledgements}

This work is supported by the National Natural Science Foundation of China (No. U24A20335, No. 62476150) and Beijing Natural Science Foundation (L243006).


\bibliography{custom}

\begin{thebibliography}{51}
\providecommand{\natexlab}[1]{#1}

\bibitem[{Achiam et~al.(2023)Achiam, Adler, Agarwal, Ahmad, Akkaya, Aleman, Almeida, Altenschmidt, Altman, Anadkat et~al.}]{achiam2023gpt}
Josh Achiam, Steven Adler, Sandhini Agarwal, Lama Ahmad, Ilge Akkaya, Florencia~Leoni Aleman, Diogo Almeida, Janko Altenschmidt, Sam Altman, Shyamal Anadkat, et~al. 2023.
\newblock Gpt-4 technical report.
\newblock \emph{arXiv preprint arXiv:2303.08774}.

\bibitem[{Bordt et~al.(2024)Bordt, Nori, and Caruana}]{bordt2024elephants}
Sebastian Bordt, Harsha Nori, and Rich Caruana. 2024.
\newblock Elephants never forget: Testing language models for memorization of tabular data.
\newblock \emph{arXiv preprint arXiv:2403.06644}.

\bibitem[{Brown(2020)}]{brown2020language}
Tom~B Brown. 2020.
\newblock Language models are few-shot learners.
\newblock \emph{arXiv preprint arXiv:2005.14165}.

\bibitem[{Chang et~al.(2024)Chang, Wang, Wang, Wu, Yang, Zhu, Chen, Yi, Wang, Wang et~al.}]{chang2024survey}
Yupeng Chang, Xu~Wang, Jindong Wang, Yuan Wu, Linyi Yang, Kaijie Zhu, Hao Chen, Xiaoyuan Yi, Cunxiang Wang, Yidong Wang, et~al. 2024.
\newblock A survey on evaluation of large language models.
\newblock \emph{ACM Transactions on Intelligent Systems and Technology}, 15(3):1--45.

\bibitem[{Chen et~al.(2024)Chen, Wang, Yao, Bai, Hou, and Li}]{chen2024finding}
Jianhui Chen, Xiaozhi Wang, Zijun Yao, Yushi Bai, Lei Hou, and Juanzi Li. 2024.
\newblock Finding safety neurons in large language models.
\newblock \emph{arXiv preprint arXiv:2406.14144}.

\bibitem[{Clark et~al.(2018)Clark, Cowhey, Etzioni, Khot, Sabharwal, Schoenick, and Tafjord}]{allenai:arc}
Peter Clark, Isaac Cowhey, Oren Etzioni, Tushar Khot, Ashish Sabharwal, Carissa Schoenick, and Oyvind Tafjord. 2018.
\newblock Think you have solved question answering? try arc, the ai2 reasoning challenge.
\newblock \emph{arXiv:1803.05457v1}.

\bibitem[{Cobbe et~al.(2021)Cobbe, Kosaraju, Bavarian, Chen, Jun, Kaiser, Plappert, Tworek, Hilton, Nakano et~al.}]{cobbe2021training}
Karl Cobbe, Vineet Kosaraju, Mohammad Bavarian, Mark Chen, Heewoo Jun, Lukasz Kaiser, Matthias Plappert, Jerry Tworek, Jacob Hilton, Reiichiro Nakano, et~al. 2021.
\newblock Training verifiers to solve math word problems.
\newblock \emph{arXiv preprint arXiv:2110.14168}.

\bibitem[{Dai et~al.(2021)Dai, Dong, Hao, Sui, Chang, and Wei}]{dai2021knowledge}
Damai Dai, Li~Dong, Yaru Hao, Zhifang Sui, Baobao Chang, and Furu Wei. 2021.
\newblock Knowledge neurons in pretrained transformers.
\newblock \emph{arXiv preprint arXiv:2104.08696}.

\bibitem[{Dekoninck et~al.(2024)Dekoninck, M{\"u}ller, Baader, Fischer, and Vechev}]{dekoninck2024evading}
Jasper Dekoninck, Mark~Niklas M{\"u}ller, Maximilian Baader, Marc Fischer, and Martin Vechev. 2024.
\newblock Evading data contamination detection for language models is (too) easy.
\newblock \emph{arXiv preprint arXiv:2402.02823}.

\bibitem[{Dong et~al.(2024)Dong, Jiang, Liu, Jin, Gu, Yang, and Li}]{dong2024generalization}
Yihong Dong, Xue Jiang, Huanyu Liu, Zhi Jin, Bin Gu, Mengfei Yang, and Ge~Li. 2024.
\newblock Generalization or memorization: Data contamination and trustworthy evaluation for large language models.
\newblock \emph{arXiv preprint arXiv:2402.15938}.

\bibitem[{Geva et~al.(2022)Geva, Caciularu, Wang, and Goldberg}]{geva2022transformer}
Mor Geva, Avi Caciularu, Kevin~Ro Wang, and Yoav Goldberg. 2022.
\newblock Transformer feed-forward layers build predictions by promoting concepts in the vocabulary space.
\newblock \emph{arXiv preprint arXiv:2203.14680}.

\bibitem[{Geva et~al.(2020)Geva, Schuster, Berant, and Levy}]{geva2020transformer}
Mor Geva, Roei Schuster, Jonathan Berant, and Omer Levy. 2020.
\newblock Transformer feed-forward layers are key-value memories.
\newblock \emph{arXiv preprint arXiv:2012.14913}.

\bibitem[{Ghandeharioun et~al.(2024)Ghandeharioun, Caciularu, Pearce, Dixon, and Geva}]{ghandeharioun2024patchscope}
Asma Ghandeharioun, Avi Caciularu, Adam Pearce, Lucas Dixon, and Mor Geva. 2024.
\newblock Patchscope: A unifying framework for inspecting hidden representations of language models.
\newblock \emph{arXiv preprint arXiv:2401.06102}.

\bibitem[{Golchin and Surdeanu(2023)}]{golchin2023time}
Shahriar Golchin and Mihai Surdeanu. 2023.
\newblock Time travel in llms: Tracing data contamination in large language models.
\newblock \emph{arXiv preprint arXiv:2308.08493}.

\bibitem[{Gou et~al.(2023)Gou, Shao, Gong, Shen, Yang, Huang, Duan, and Chen}]{gou2023tora}
Zhibin Gou, Zhihong Shao, Yeyun Gong, Yelong Shen, Yujiu Yang, Minlie Huang, Nan Duan, and Weizhu Chen. 2023.
\newblock Tora: A tool-integrated reasoning agent for mathematical problem solving.
\newblock \emph{arXiv preprint arXiv:2309.17452}.

\bibitem[{Guo et~al.(2023)Guo, Jin, Liu, Huang, Shi, Yu, Liu, Li, Xiong, Xiong et~al.}]{guo2023evaluating}
Zishan Guo, Renren Jin, Chuang Liu, Yufei Huang, Dan Shi, Linhao Yu, Yan Liu, Jiaxuan Li, Bojian Xiong, Deyi Xiong, et~al. 2023.
\newblock Evaluating large language models: A comprehensive survey.
\newblock \emph{arXiv preprint arXiv:2310.19736}.

\bibitem[{Gurnee et~al.(2024)Gurnee, Horsley, Guo, Kheirkhah, Sun, Hathaway, Nanda, and Bertsimas}]{gurnee2024universal}
Wes Gurnee, Theo Horsley, Zifan~Carl Guo, Tara~Rezaei Kheirkhah, Qinyi Sun, Will Hathaway, Neel Nanda, and Dimitris Bertsimas. 2024.
\newblock Universal neurons in gpt2 language models.
\newblock \emph{arXiv preprint arXiv:2401.12181}.

\bibitem[{Hendrycks et~al.(2020)Hendrycks, Burns, Basart, Zou, Mazeika, Song, and Steinhardt}]{hendrycks2020measuring}
Dan Hendrycks, Collin Burns, Steven Basart, Andy Zou, Mantas Mazeika, Dawn Song, and Jacob Steinhardt. 2020.
\newblock Measuring massive multitask language understanding.
\newblock \emph{arXiv preprint arXiv:2009.03300}.

\bibitem[{Hendrycks et~al.(2021)Hendrycks, Burns, Kadavath, Arora, Basart, Tang, Song, and Steinhardt}]{hendrycks2021measuring}
Dan Hendrycks, Collin Burns, Saurav Kadavath, Akul Arora, Steven Basart, Eric Tang, Dawn Song, and Jacob Steinhardt. 2021.
\newblock Measuring mathematical problem solving with the math dataset.
\newblock \emph{arXiv preprint arXiv:2103.03874}.

\bibitem[{Jacovi et~al.(2023)Jacovi, Caciularu, Goldman, and Goldberg}]{jacovi2023stop}
Alon Jacovi, Avi Caciularu, Omer Goldman, and Yoav Goldberg. 2023.
\newblock Stop uploading test data in plain text: Practical strategies for mitigating data contamination by evaluation benchmarks.
\newblock \emph{arXiv preprint arXiv:2305.10160}.

\bibitem[{Jiang et~al.(2023)Jiang, Sablayrolles, Mensch, Bamford, Chaplot, Casas, Bressand, Lengyel, Lample, Saulnier et~al.}]{jiang2023mistral}
Albert~Q Jiang, Alexandre Sablayrolles, Arthur Mensch, Chris Bamford, Devendra~Singh Chaplot, Diego de~las Casas, Florian Bressand, Gianna Lengyel, Guillaume Lample, Lucile Saulnier, et~al. 2023.
\newblock Mistral 7b.
\newblock \emph{arXiv preprint arXiv:2310.06825}.

\bibitem[{Koncel-Kedziorski et~al.(2016)Koncel-Kedziorski, Roy, Amini, Kushman, and Hajishirzi}]{koncel2016mawps}
Rik Koncel-Kedziorski, Subhro Roy, Aida Amini, Nate Kushman, and Hannaneh Hajishirzi. 2016.
\newblock Mawps: A math word problem repository.
\newblock In \emph{Proceedings of the 2016 conference of the north american chapter of the association for computational linguistics: human language technologies}, pages 1152--1157.

\bibitem[{Li(2023)}]{li2023estimating}
Yucheng Li. 2023.
\newblock Estimating contamination via perplexity: Quantifying memorisation in language model evaluation.
\newblock \emph{arXiv preprint arXiv:2309.10677}.

\bibitem[{Li et~al.(2024{\natexlab{a}})Li, Guerin, and Lin}]{li2024latesteval}
Yucheng Li, Frank Guerin, and Chenghua Lin. 2024{\natexlab{a}}.
\newblock Latesteval: Addressing data contamination in language model evaluation through dynamic and time-sensitive test construction.
\newblock In \emph{Proceedings of the AAAI Conference on Artificial Intelligence}, volume~38, pages 18600--18607.

\bibitem[{Li et~al.(2024{\natexlab{b}})Li, Guo, Guerin, and Lin}]{li2024open}
Yucheng Li, Yunhao Guo, Frank Guerin, and Chenghua Lin. 2024{\natexlab{b}}.
\newblock An open-source data contamination report for large language models.
\newblock In \emph{Findings of the Association for Computational Linguistics: EMNLP 2024}, pages 528--541.

\bibitem[{Lian et~al.(2023)Lian, Goodson, Pentland et~al.}]{lian2023openorca}
W~Lian, B~Goodson, E~Pentland, et~al. 2023.
\newblock Openorca: An open dataset of gpt augmented flan reasoning traces.

\bibitem[{Litschko et~al.(2023)Litschko, M{\"u}ller-Eberstein, Van Der~Goot, Weber, and Plank}]{litschko2023establishing}
Robert Litschko, Max M{\"u}ller-Eberstein, Rob Van Der~Goot, Leon Weber, and Barbara Plank. 2023.
\newblock Establishing trustworthiness: Rethinking tasks and model evaluation.
\newblock \emph{arXiv preprint arXiv:2310.05442}.

\bibitem[{Magar and Schwartz(2022)}]{magar2022data}
Inbal Magar and Roy Schwartz. 2022.
\newblock Data contamination: From memorization to exploitation.
\newblock \emph{arXiv preprint arXiv:2203.08242}.

\bibitem[{Matton et~al.(2024)Matton, Sherborne, Aumiller, Tommasone, Alizadeh, He, Ma, Voisin, Gilsenan-McMahon, and Gall{\'e}}]{matton2024leakage}
Alexandre Matton, Tom Sherborne, Dennis Aumiller, Elena Tommasone, Milad Alizadeh, Jingyi He, Raymond Ma, Maxime Voisin, Ellen Gilsenan-McMahon, and Matthias Gall{\'e}. 2024.
\newblock On leakage of code generation evaluation datasets.
\newblock \emph{arXiv preprint arXiv:2407.07565}.

\bibitem[{Meng et~al.(2022)Meng, Bau, Andonian, and Belinkov}]{meng2022locating}
Kevin Meng, David Bau, Alex Andonian, and Yonatan Belinkov. 2022.
\newblock Locating and editing factual associations in gpt.
\newblock \emph{Advances in Neural Information Processing Systems}, 35:17359--17372.

\bibitem[{Miao et~al.(2021)Miao, Liang, and Su}]{miao2021diverse}
Shen-Yun Miao, Chao-Chun Liang, and Keh-Yih Su. 2021.
\newblock A diverse corpus for evaluating and developing english math word problem solvers.
\newblock \emph{arXiv preprint arXiv:2106.15772}.

\bibitem[{Ni et~al.(2024)Ni, Xue, Yue, Deng, Shah, Jain, Neubig, and You}]{ni2024mixeval}
Jinjie Ni, Fuzhao Xue, Xiang Yue, Yuntian Deng, Mahir Shah, Kabir Jain, Graham Neubig, and Yang You. 2024.
\newblock Mixeval: Deriving wisdom of the crowd from llm benchmark mixtures.
\newblock \emph{arXiv preprint arXiv:2406.06565}.

\bibitem[{Patel et~al.(2021)Patel, Bhattamishra, and Goyal}]{patel2021nlp}
Arkil Patel, Satwik Bhattamishra, and Navin Goyal. 2021.
\newblock Are nlp models really able to solve simple math word problems?
\newblock \emph{arXiv preprint arXiv:2103.07191}.

\bibitem[{Sainz et~al.(2023)Sainz, Campos, Garc{\'\i}a-Ferrero, Etxaniz, de~Lacalle, and Agirre}]{sainz2023nlp}
Oscar Sainz, Jon~Ander Campos, Iker Garc{\'\i}a-Ferrero, Julen Etxaniz, Oier~Lopez de~Lacalle, and Eneko Agirre. 2023.
\newblock Nlp evaluation in trouble: On the need to measure llm data contamination for each benchmark.
\newblock \emph{arXiv preprint arXiv:2310.18018}.

\bibitem[{Toshniwal et~al.(2024)Toshniwal, Du, Moshkov, Kisacanin, Ayrapetyan, and Gitman}]{toshniwal2024openmathinstruct}
Shubham Toshniwal, Wei Du, Ivan Moshkov, Branislav Kisacanin, Alexan Ayrapetyan, and Igor Gitman. 2024.
\newblock Openmathinstruct-2: Accelerating ai for math with massive open-source instruction data.
\newblock \emph{arXiv preprint arXiv:2410.01560}.

\bibitem[{Touvron et~al.(2023)Touvron, Lavril, Izacard, Martinet, Lachaux, Lacroix, Rozi{\`e}re, Goyal, Hambro, Azhar et~al.}]{touvron2023llama}
Hugo Touvron, Thibaut Lavril, Gautier Izacard, Xavier Martinet, Marie-Anne Lachaux, Timoth{\'e}e Lacroix, Baptiste Rozi{\`e}re, Naman Goyal, Eric Hambro, Faisal Azhar, et~al. 2023.
\newblock Llama: Open and efficient foundation language models.
\newblock \emph{arXiv preprint arXiv:2302.13971}.

\bibitem[{Tu et~al.(2024)Tu, Zhu, Bai, Yao, Hou, and Li}]{tu2024dice}
Shangqing Tu, Kejian Zhu, Yushi Bai, Zijun Yao, Lei Hou, and Juanzi Li. 2024.
\newblock Dice: Detecting in-distribution contamination in llm's fine-tuning phase for math reasoning.
\newblock \emph{arXiv preprint arXiv:2406.04197}.

\bibitem[{Vaswani(2017)}]{vaswani2017attention}
A~Vaswani. 2017.
\newblock Attention is all you need.
\newblock \emph{Advances in Neural Information Processing Systems}.

\bibitem[{Vig et~al.(2020{\natexlab{a}})Vig, Gehrmann, Belinkov, Qian, Nevo, Singer, and Shieber}]{vig2020investigating}
Jesse Vig, Sebastian Gehrmann, Yonatan Belinkov, Sharon Qian, Daniel Nevo, Yaron Singer, and Stuart Shieber. 2020{\natexlab{a}}.
\newblock Investigating gender bias in language models using causal mediation analysis.
\newblock \emph{Advances in neural information processing systems}, 33:12388--12401.

\bibitem[{Vig et~al.(2020{\natexlab{b}})Vig, Gehrmann, Belinkov, Qian, Nevo, Singer, and Shieber}]{Vig_Gehrmann_Belinkov_Qian_Nevo_Singer_Shieber_2020}
Jesse Vig, Sebastian Gehrmann, Yonatan Belinkov, Sharon Qian, Daniel Nevo, Yoram Singer, and StuartM. Shieber. 2020{\natexlab{b}}.
\newblock Investigating gender bias in language models using causal mediation analysis.
\newblock \emph{Neural Information Processing Systems,Neural Information Processing Systems}.

\bibitem[{Wang et~al.(2023)Wang, Yang, Huang, Yang, Majumder, and Wei}]{wang2023improving}
Liang Wang, Nan Yang, Xiaolong Huang, Linjun Yang, Rangan Majumder, and Furu Wei. 2023.
\newblock Improving text embeddings with large language models.
\newblock \emph{arXiv preprint arXiv:2401.00368}.

\bibitem[{Wang et~al.(2022)Wang, Wen, Zhang, Hou, Liu, and Li}]{wang2022finding}
Xiaozhi Wang, Kaiyue Wen, Zhengyan Zhang, Lei Hou, Zhiyuan Liu, and Juanzi Li. 2022.
\newblock Finding skill neurons in pre-trained transformer-based language models.
\newblock \emph{arXiv preprint arXiv:2211.07349}.

\bibitem[{Xu et~al.(2024)Xu, Wang, Fan, and Liu}]{xu2024benchmarking}
Ruijie Xu, Zengzhi Wang, Run-Ze Fan, and Pengfei Liu. 2024.
\newblock Benchmarking benchmark leakage in large language models.
\newblock \emph{arXiv preprint arXiv:2404.18824}.

\bibitem[{Yu et~al.(2023)Yu, Wang, Tu, Cao, Zhang-Li, Lv, Peng, Yao, Zhang, Li et~al.}]{yu2023kola}
Jifan Yu, Xiaozhi Wang, Shangqing Tu, Shulin Cao, Daniel Zhang-Li, Xin Lv, Hao Peng, Zijun Yao, Xiaohan Zhang, Hanming Li, et~al. 2023.
\newblock Kola: Carefully benchmarking world knowledge of large language models.
\newblock \emph{arXiv preprint arXiv:2306.09296}.

\bibitem[{Yu et~al.(2024)Yu, Gao, Yao, Wang, Ye, Wang, Xie, Zhang, and Zhang}]{yu2024kieval}
Zhuohao Yu, Chang Gao, Wenjin Yao, Yidong Wang, Wei Ye, Jindong Wang, Xing Xie, Yue Zhang, and Shikun Zhang. 2024.
\newblock Kieval: A knowledge-grounded interactive evaluation framework for large language models.
\newblock \emph{arXiv preprint arXiv:2402.15043}.

\bibitem[{Zhang and Nanda(2023)}]{zhang2023towards}
Fred Zhang and Neel Nanda. 2023.
\newblock Towards best practices of activation patching in language models: Metrics and methods.
\newblock \emph{arXiv preprint arXiv:2309.16042}.

\bibitem[{Zhao et~al.(2024)Zhao, Chen, Yang, Liu, Deng, Cai, Wang, Yin, and Du}]{zhao2024explainability}
Haiyan Zhao, Hanjie Chen, Fan Yang, Ninghao Liu, Huiqi Deng, Hengyi Cai, Shuaiqiang Wang, Dawei Yin, and Mengnan Du. 2024.
\newblock Explainability for large language models: A survey.
\newblock \emph{ACM Transactions on Intelligent Systems and Technology}, 15(2):1--38.

\bibitem[{Zhao et~al.(2023)Zhao, Zhou, Li, Tang, Wang, Hou, Min, Zhang, Zhang, Dong et~al.}]{zhao2023survey}
Wayne~Xin Zhao, Kun Zhou, Junyi Li, Tianyi Tang, Xiaolei Wang, Yupeng Hou, Yingqian Min, Beichen Zhang, Junjie Zhang, Zican Dong, et~al. 2023.
\newblock A survey of large language models.
\newblock \emph{arXiv preprint arXiv:2303.18223}.

\bibitem[{Zhou et~al.(2023)Zhou, Zhu, Chen, Chen, Zhao, Chen, Lin, Wen, and Han}]{zhou2023don}
Kun Zhou, Yutao Zhu, Zhipeng Chen, Wentong Chen, Wayne~Xin Zhao, Xu~Chen, Yankai Lin, Ji-Rong Wen, and Jiawei Han. 2023.
\newblock Don't make your llm an evaluation benchmark cheater.
\newblock \emph{arXiv preprint arXiv:2311.01964}.

\bibitem[{Zhu et~al.(2023{\natexlab{a}})Zhu, Chen, Wang, Zhenqiang~Gong, Yang, and Xie}]{zhu2023dyval}
Kaijie Zhu, Jiaao Chen, Jindong Wang, Neil Zhenqiang~Gong, Diyi Yang, and Xing Xie. 2023{\natexlab{a}}.
\newblock Dyval: Graph-informed dynamic evaluation of large language models.
\newblock \emph{arXiv e-prints}, pages arXiv--2309.

\bibitem[{Zhu et~al.(2023{\natexlab{b}})Zhu, Hao, He, Song, Zhang, Hu, Wei, Wang, and Lu}]{zhu2023clean}
Wenhong Zhu, Hongkun Hao, Zhiwei He, Yunze Song, Yumeng Zhang, Hanxu Hu, Yiran Wei, Rui Wang, and Hongyuan Lu. 2023{\natexlab{b}}.
\newblock Clean-eval: Clean evaluation on contaminated large language models.
\newblock \emph{arXiv preprint arXiv:2311.09154}.

\end{thebibliography}
\clearpage
\appendix

\section{Real World Application}
\label{appendix: real-world}
As shown in Table \ref{tab:whole real-world results}, we select a range of models available on Hugging Face, which we applied our method to. We calculated the original scores on GSM8K (Zero-Shot), as well as scores under our evaluation framework. The deeper the orange in cell \textbf{$\Delta_\text{acc}$}, the more severe the contamination present in the model. It was observed that the accuracy of llamaRAGdrama and Fewshot-Metamath-OrcaVicuna-Mistral experienced a significant decline, suggesting that both may have serious contamination on the GSM8K dataset or have obtained input shortcuts by fitting the I/O format of GSM8K.

\section{Cost of Our Method}

As shown in Table \ref{cost}, our expenses are solely on training cards, which are significantly lower than the labor and computational costs associated with maintaining the dynamic benchmark.

\section{Generalization}\label{appendix_dataset_construct}

In this section, we will introduce various contamination scenarios for proving generalizability.

\subsection{Different Benchmarks}

Similar to the GSM8K dataset in main experiments as shown in Table \ref{trustworthy evaluation results}, we fully fine-tuned a series of contaminated and uncontaminated models on different benchmark (e.g. MAWPS, MATH ), as shown in Table \ref{appendix:generalization on benchmarks}. The results of MAWPS are shown in Table \ref{tab:generalization} in the main text, and the results of MATH are shown in Table \ref{appendix:MATH} in the Appendix.

\begin{table}
  \setlength\tabcolsep{4pt}
\begin{tabular}{lccc}
\toprule
 & \multicolumn{3}{c}{\textbf{MATH}} \\
 \cmidrule(lr){2-4} 
 & \textbf{Ori.} & \textbf{TE} & \textbf{Ref.} \\
\midrule
 Vanilla & 8.5 & 8.5 & 16.7 \\
~+MATH & 16.8 & \cellcolor[HTML]{FDB594} 11.0(\textcolor{darkred}{-5.8}) & 25.7 \\
~+MATH-Syn & 13.9 & \cellcolor[HTML]{FED9C7} 10.5(\textcolor{darkred}{-3.4}) & 21.3 \\
~+5$\times$MATH & 29.6 & \cellcolor[HTML]{FC8D59} 12.8(\textcolor{darkred}{-16.8}) & 28.5 \\
~+5$\times$MATH-Syn & 19.5 & \cellcolor[HTML]{FDB594} 9.5(\textcolor{darkred}{-10.0}) & 23.1 \\
      \midrule
~+OpenOrca & 11.5 & \cellcolor[HTML]{FEF6F1} 11.0(\textcolor{darkred}{-0.5}) & 21.0\\
~+SVAMP & 11.0 & 11.0(\textcolor{darkblue}{+0.0}) & 26.2 \\
~+ASDiv & 10.9 & \cellcolor[HTML]{C3DCEC} 11.5(\textcolor{darkblue}{+0.6}) & 23.5 \\

\bottomrule
\end{tabular}
\caption{\label{appendix:MATH}
    Generalization of our method on  MATH.
  }
\end{table}

\begin{table}
  \setlength\tabcolsep{4pt}
\begin{tabular}{lcccc}
\toprule
 & \multicolumn{4}{c}{\textbf{LLaMA-3-8B-Instruct}} \\
 \cmidrule(lr){2-5} 
 & \textbf{Ref Acc} & \textbf{Ori.} & \textbf{TE} & \textbf{$ \Delta_\text{acc} $}  \\
\midrule
 Vanilla & 67.0 & 61.0 & 61.0 & -  \\
~+GSM-i & 69.3 & 77.9 & 65.6 & \cellcolor[HTML]{FDB594} \textcolor{darkred}{-12.3}  \\
~+GSM-i-Syn & 66.8 & 71.1 & 62.6 & \cellcolor[HTML]{FED9C7} \textcolor{darkred}{-8.5}  \\
~+5$\times$GSM-i & 70.1 & 90.0 & 67.9 & \cellcolor[HTML]{FC8D59} \textcolor{darkred}{-22.1}  \\
~+5$\times$GSM-i-Syn & 67.4 & 74.3 & 63.8 & \cellcolor[HTML]{FCAB84} \textcolor{darkred}{-10.5} \\
      \midrule
~+OpenOrca & 68.2 & 58.9 & 61.3 & \cellcolor[HTML]{C3DCEC}\textcolor{darkblue}{+2.4} \\
~+GSM8K Train & 67.5 & 66.2 & 61.6 & \cellcolor[HTML]{FED9C7}\textcolor{darkred}{-4.6} \\
~+MATH & 65.7 & 59.4 & 59.5 & \cellcolor[HTML]{DCEBF4}\textcolor{darkblue}{+0.1} \\
~+MATH-Syn & 66.8 & 60.6 & 59.8 & \cellcolor[HTML]{FEF6F1}\textcolor{darkred}{-0.8} \\

\bottomrule
\end{tabular}
\caption{\label{Other Architectures}
    The generalizability of our evaluation method to different architectures.
  }
\end{table}


\subsection{Various Hyperparameters}

To evaluate the robustness of our method, we varied several training strategies (learning rate), as well as the frequency of contaminated samples. As shown in Figure \ref{fig:generalization} of the main text, we present the evaluation results of a series of contaminated and uncontaminated models under these varying settings using our method, demonstrating a strong correlation with MixEval. Below, we provide a detailed description of the settings used in this study, as summarized in Table \ref{exp:ablation-models}. By modifying different training settings, we generated a total of 72 models for evaluation, as detailed below:

\begin{enumerate}
    \item \textbf{Datasets.} Using GSM8K as the benchmark to be tested, we fine-tuned the model using GSM8K and a series of OOD datasets.

    \item \textbf{Occurrences.} Between 1 and 20 times.

    \item \textbf{Learning Rate.} Select different learning rates for various training methods.

\end{enumerate}

\subsection{Different Architecture}

Furthermore, we evaluated the effectiveness of our method when applied to the LLaMA3-8B architecture. As shown in Table \ref{trustworthy evaluation results} of the main text, we simulated several \textit{contaminated} and \textit{uncontaminated} models on the GSM8K dataset through supervised fine-tuning (SFT). Using our method, we successfully identified a new set of shortcut neurons within the LLaMA3-8B architecture. We then applied our evaluation approach to these models, with the results presented in Table \ref{Other Architectures}. Our method demonstrated good performance on LLaMA3-8B, effectively reducing the performance of contaminated models to normal levels while preserving the original performance of uncontaminated models.


\begin{table*}
    \centering
  \fontsize{11pt}{15pt}\selectfont {
	\begin{tabular}{ccccc}
		\toprule
		\multicolumn{3}{c}{\textbf{Locating (Per Arch.)}} & \multicolumn{2}{c}{\textbf{Evaluation}} \\ 
         \cmidrule(lr){1-3}    \cmidrule(lr){4-5}     
		\textbf{Comparative} &\textbf{Causality} &\textbf{GPU}& \textbf{Time (Per Batch)} & \textbf{GPU}\\ 
        \hline
         6h &72h & 3$\times A100$& 10s & 2$\times A100$\\
		\bottomrule
	\end{tabular}
	}
        \caption{The cost of our evaluation method for one 7B model architecture. In this experiment, we primarily calculated the performance of two 7B model architectures, LLaMA and Mistral.}
        \label{cost}
\end{table*}

\begin{table*}
  \centering
  \fontsize{9pt}{12pt}\selectfont 
  \begin{tabular}{lcccc}
    \hline
    \textbf{Label}           & \textbf{Benchmark Samples} & \textbf{Occurrences} & \textbf{Learning Rate} & \textbf{Base Models} \\
    \hline
    contaminated       & \{GSM-i, GSM-i-Syn\}           &  \{1,5,10,15,20\}       & \multirow{2}{*}{\{1e-3, 1e-5, 1e-8\}}  & \multirow{2}{*}{\{LLaMA2-7B, Mistral-7B-v0.2\}}                 \\
    uncontaminated     & \{GSM8K Train, MATH(-Syn)\}          &   \{1\}                        \\
    \hline
  \end{tabular}
  \caption{\label{exp:ablation-models}
    The models needed in the trustworthy evaluation experiment are all fine-tuned from the given basic models, simulating a variety of contaminated and uncontaminated models in the real world.
  }
\end{table*}

\begin{table*}
  \centering
  \setlength\tabcolsep{5pt}
\begin{tabular}{lcccc}
\toprule
\textbf{Models} & \textbf{Ref.} & \textbf{Ori.} & \textbf{TE}  & \textbf{$\Delta_\text{acc}$} \\
\midrule
llamaRAGdrama & 15.5 & 45.2 & 21.7 & \cellcolor[HTML]{FC8D59}  \textcolor{darkred}{-23.5}  \\
Metamath-reproduce-7b & 59.2 & 64.0 & 59.0 & \cellcolor[HTML]{FEECE4} \textcolor{darkred}{-5.0} \\
Llama-2-7b-gsm8k & 36.6 & 34.0 & 34.5 & \cellcolor[HTML]{E7F1F7} \textcolor{darkblue}{+0.5} \\
llemma\_7b & 24.6 & 27.5 & 29.3 & \cellcolor[HTML]{B8D5E8} \textcolor{darkblue}{+1.8}  \\
StableBeluga-7B-activity-fine-tuned-v2 & 18.3 & 19.0 & 20.5 & \cellcolor[HTML]{C3DCEC}\textcolor{darkblue}{+1.5} \\
Llama-2-7b-chat-hf-20-sparsity & 15.5 & 18.6 & 18.1 & 
\cellcolor[HTML]{FEF6F1} \textcolor{darkred}{-0.5}  \\
\midrule
Calme-7B-Instruct-v0.4 & 67.6 & 75.3 & 65.8 & \cellcolor[HTML]{FEE2D6} \textcolor{darkred}{-9.5}\\ 
flux-7b-v0.2 & 70.5 & 71.6 & 73.3 & \cellcolor[HTML]{E7F1F7} \textcolor{darkblue}{+1.7} \\
mistral-ft-optimized-1218 & 70.1 & 73.4 & 68.6 & \cellcolor[HTML]{FEECE4} \textcolor{darkred}{-4.8} \\
ladybird-base-7B-v8 & 57.0 & 63.5 &  65.7 & \cellcolor[HTML]{C5DDEC} \textcolor{darkblue}{+2.2} \\
Fewshot-Metamath-OrcaVicuna-Mistral & 57.5 & 66.4  & 50.1 & \cellcolor[HTML]{FC8D59} \textcolor{darkred}{-16.3} \\
MetaMath-Mistral-7B & 65.5 & 70.8 &  67.6 & \cellcolor[HTML]{FEECE4} \textcolor{darkred}{-3.2} \\
openchat-nectar-0.1 & 49.3 & 63.3 & 51.6 & \cellcolor[HTML]{FDC8AF} \textcolor{darkred}{-11.7} \\
K2S3-Mistral-7b-v1.2 & 44.4 & 53.9 & 51.8 & \cellcolor[HTML]{FEF3ED} \textcolor{darkred}{-2.1} \\
TopicNeuralHermes-2.5-Mistral-7B & 52.1 & 54.5 & 56.7 & 
\cellcolor[HTML]{B8D5E8} \textcolor{darkblue}{+2.2} \\
mistral-maths7B & 47.9 & 43.5 & 47.6 & \cellcolor[HTML]{A1C8E0} \textcolor{darkblue}{+4.1} \\
mistralv1\_gsm8k\_merged & 40.8 & 49.7 & 41.0 & \cellcolor[HTML]{FEDBCB}\textcolor{darkred}{-8.7} \\
Hyperion-3.0-Mistral-7B-DPO & 42.2 & 44.9 & 45.5 & \cellcolor[HTML]{FEF6F1} \textcolor{darkred}{-0.6} \\
Hercules-3.1-Mistral-7B & 43.7 & 43.0 & 44.8 & \cellcolor[HTML]{E7F1F7} \textcolor{darkblue}{+1.8} \\
\bottomrule
\end{tabular}
\caption{\label{tab:whole real-world results}
    Real-world models with LLaMA and Mistral architecture are downloaded from huggingface. Ref. is the score calculateed on MixEval, which is a relatively fair score. The number of \textbf{$\Delta_\text{acc}$} represents TE minus Ori.
  }
\end{table*}

\begin{table*}
  \centering
  \fontsize{10pt}{12pt}\selectfont 
  \begin{tabular}{lccc}
    \hline
    \textbf{Label}           & \textbf{Benchmark Samples} & \textbf{Occurrences} & \textbf{Base Models} \\
    \hline
    contaminated       & \{$\mathcal{D}$, $\mathcal{D}$-Syn\}           &  \{1,5\}       & \multirow{2}{*}{\{LLaMA2-7B, Mistral-7B-v0.2\}}                    \\
    uncontaminated     & \{SVAMP, ASDiv\}          &   \{1\}            \\
    \hline
  \end{tabular}
  \caption{\label{appendix:generalization on benchmarks}
    The settings for contaminated and uncontaminated models when the benchmark is $\mathcal{D}$ (e.g. MATH, MAWPS). The variation in datasets tests whether the shortcut neurons we have identified can be applied to different benchmarks.
  }
\end{table*}

\begin{table*}
  \centering
  \setlength\tabcolsep{5pt}
\begin{tabular}{lcccccccc}
\toprule
 & \multicolumn{4}{c}{\textbf{LLaMA2-7B}} & \multicolumn{4}{c}{\textbf{Mistral-7B}} \\
 \cmidrule(lr){2-5} \cmidrule(lr){6-9}
 & \textbf{Ref Acc} & \textbf{Ori.} & \textbf{TE} & \textbf{$ \Delta_\text{acc} $} & \textbf{Ref Acc} & \textbf{Ori.} & \textbf{TE} & \textbf{$\Delta_\text{acc}$} \\
\midrule
 Vanilla & 16.7 & 18.5 & 18.5 & - & 31.8 & 40.0 & 40.0 & - \\
~+GSM-i-r & 25.3 & 41.6 & 27.9 & \cellcolor[HTML]{FDB594} \textcolor{darkred}{-13.7} & 37.3 & 60.7 & 41.1 & \cellcolor[HTML]{FCAB84} \textcolor{darkred}{-19.6} \\
~+GSM-i-Syn-r & 22.5 & 34.6 & 21.8 & \cellcolor[HTML]{FED9C7} \textcolor{darkred}{-12.8} & 36.5 & 47.3 & 39.4 & \cellcolor[HTML]{FEE5D9}\textcolor{darkred}{-7.9} \\
~+5$\times$GSM-i-r & 26.9 & 77.2 & 29.8 & \cellcolor[HTML]{FC8D59} \textcolor{darkred}{-47.4} & 40.2 & 87.1 & 48.6 & \cellcolor[HTML]{FC8D59} \textcolor{darkred}{-38.5} \\
~+5$\times$GSM-i-Syn-r & 23.1 & 42.6 & 22.8 & \cellcolor[HTML]{FCAB84} \textcolor{darkred}{-19.8} & 37.9 & 55.7 & 42.8 & \cellcolor[HTML]{FDB594} \textcolor{darkred}{-12.9} \\

\bottomrule
\end{tabular}
\caption{\label{random order}
    The generalizability of our evaluation method to the order in which contaminated samples appear. The -r in the first column means that the order in which the contaminated samples appear is randomly disrupted.
  }
\end{table*}

\begin{table*}
  \centering
  \setlength\tabcolsep{5pt}
\begin{tabular}{lcccccccccc}
\toprule
& \multicolumn{1}{c}{} & \multicolumn{2}{c}{\textbf{Our Method}} & \multicolumn{6}{c}{\textbf{KIEval}} \\
 \cmidrule(lr){3-4}  \cmidrule(lr){5-10}
 & \textbf{Ori.(5-shot)} & \textbf{TE}  &  \textbf{$ \Delta_\text{acc} $}  & \textbf{Acc.} & \textbf{Log.} & \textbf{Rel.} & \textbf{Coh.} & \textbf{Con.} & \textbf{Overall}\\
\midrule
 Normal (LLaMA 2 7B + SFT) & 52.8 & 55.7 & \cellcolor[HTML]{C3DCEC}\textcolor{darkblue}{+2.9}  & 61.7 & 62.1 & 84.4 & 69.2 & 70.6 & 66.3  \\
~SFT-Cheater & 69.8 & 53.8 & \cellcolor[HTML]{FDB594} \textcolor{darkred}{-16.0} & 52.8 & 52.3 & 72.8 & 60.2 & 57.7 & 56.1 \\
~PT-Cheater & 76.8 & 59.3 & \cellcolor[HTML]{FCAB84} \textcolor{darkred}{-17.5} & 50.8 & 49.9 & 65.6 & 54.5 & 49.0 & 51.2 \\
LLaMA 2 7B Chat & 57.8 & 61.2 & \cellcolor[HTML]{A1C8E0} \textcolor{darkblue}{+3.4} & 75.3 & 75.9 & 90.1 & 80.2 & 74.0 & 77.9  \\

\bottomrule
\end{tabular}
\caption{\label{SFT and PT}
    The effect of shortcut neuron patching under two contamination strategies: SFT-Cheater (contamination via supervised fine-tuning) and PT-Cheater (contamination via continued pretraining). The test set is ARC-Challenge.
  }
\end{table*}

\subsection{Different Order of Training Data}
\label{appendix: random order}

To further evaluate the generalizability of our method, we randomized the order of contaminated samples during the SFT stage used to construct the contaminated models. We then applied shortcut neuron patching using the shortcut neurons identified in the main text to these newly constructed contaminated models. As shown in Table \ref{random order}, our method still achieved favorable trustworthy evaluation results, effectively reducing the performance of contaminated models to a normal level.

\subsection{Different Task Scenarios}

Since the experiments in the main text are all based on mathematical benchmarks, we additionally applied our method in a different task scenario. Specifically, we followed the setup of KIEval~\cite{yu2024kieval}, a recent and excellent work on trustworthy evaluation, and located a set of shortcut neurons on the ARC-Challenge dataset~\cite{allenai:arc}. We then applied our evaluation method to two types of contaminated models (both the continual pretraining phase and the SFT phase) released by KIEval and available on Hugging Face. The results, shown in Table \ref{SFT and PT}, demonstrate that our method effectively mitigates contamination effects across both SFT and continual pretraining stages, enabling fair evaluation in a different task domain.




\end{document}